\documentclass[10pt,twocolumn,letterpaper]{article}

\usepackage{wacv}
\usepackage{times}
\usepackage{epsfig}
\usepackage{graphicx}
\usepackage{amsmath}
\usepackage{amssymb}
\usepackage[ruled, lined, commentsnumbered, longend]{algorithm2e}
\usepackage{makecell}
\usepackage{footmisc}
\usepackage{algpseudocode}
\usepackage{soul}

\wacvfinalcopy 

\ifwacvfinal
\def\assignedStartPage{1} 
\fi

\ifwacvfinal
\usepackage[breaklinks=true,breaklinks=true,colorlinks,bookmarks=false]{hyperref}
\else
\usepackage[pagebackref=true,breaklinks=true,colorlinks,bookmarks=false]{hyperref}
\fi

\ifwacvfinal
\else
\fi

\begin{document}

\title{Dense-Resolution Network for Point Cloud Classification and Segmentation}


\author{Shi Qiu$^{1,2}$, Saeed Anwar$^{1,2}$ and Nick Barnes$^{1}$\\
 $^1$Australian National University, $^2$Data61-CSIRO, Australia\\
{\tt\small \{shi.qiu, saeed.anwar\}@data61.csiro.au, nick.barnes@anu.edu.au}
}

\maketitle

\begin{abstract}
Point cloud analysis is attracting attention from Artificial Intelligence research since it can be widely used in applications such as robotics, Augmented Reality, self-driving. However, it is always challenging due to irregularities, unorderedness, and sparsity. In this article, we propose a novel network named Dense-Resolution Network (DRNet) for point cloud analysis. Our DRNet is designed to learn local point features from the point cloud in different resolutions. In order to learn local point groups more effectively, we present a novel grouping method for local neighborhood searching and an error-minimizing module for capturing local features. In addition to validating the network on widely used point cloud segmentation and classification benchmarks, we also test and visualize the performance of the components. Comparing with other state-of-the-art methods, our network shows superiority on ModelNet40, ShapeNet synthetic and ScanObjectNN real point cloud datasets.
\end{abstract}

\section{Introduction}
\label{sec:intro}
With the help of rapid progress in 3D sensing technology, an increasing number of researchers are now focusing on 3D point clouds. Different from complex 3D data \eg, mesh and volumetric data, point clouds have a simpler data format. Typically, point clouds are easier to collect using different types of scanners~\cite{blais2004review} with specific algorithms: \eg, LiDAR scanners~\cite{jaboyedoff2012use} and Simultaneous localization and mapping (SLAM) algorithms. Traditional algorithms addressing point cloud learning~\cite{schnabel2007efficient,mitra2004registration,rusu2009fast,vosselman20013d} used to estimate geometric information and capture indirect clues utilizing complicated models. In contrast, deep learning models provide explicit and effective data-driven approaches to acquire information from 3D point cloud data, leveraging Convolutional Neural Networks (CNN).

In general, CNN-related methods for 3D point clouds can be divided mainly into two categories~\cite{guo2020deep}. The first one is conversion-based, which converts the 3D data to some intermediate representations, for example, MVCNN~\cite{su2015multi} projects 3D shapes into multi-view 2D images, and VoxNet~\cite{maturana2015voxnet} transfers point clouds as volumetric grids. The other one is point-based such as PointNet~\cite{qi2017pointnet}, which directly processes points. The point-based approach has become popular due to the introduction of the multi-layer perceptrons (MLPs) operation in~\cite{qi2017pointnet}. The subsequent algorithms~\cite{wang2019dynamic, Thomas_2019_ICCV} adopted MLPs to learn the local features of point clouds using graph context and kernel points.
\begin{figure}
\begin{center}
\includegraphics[width=\columnwidth]{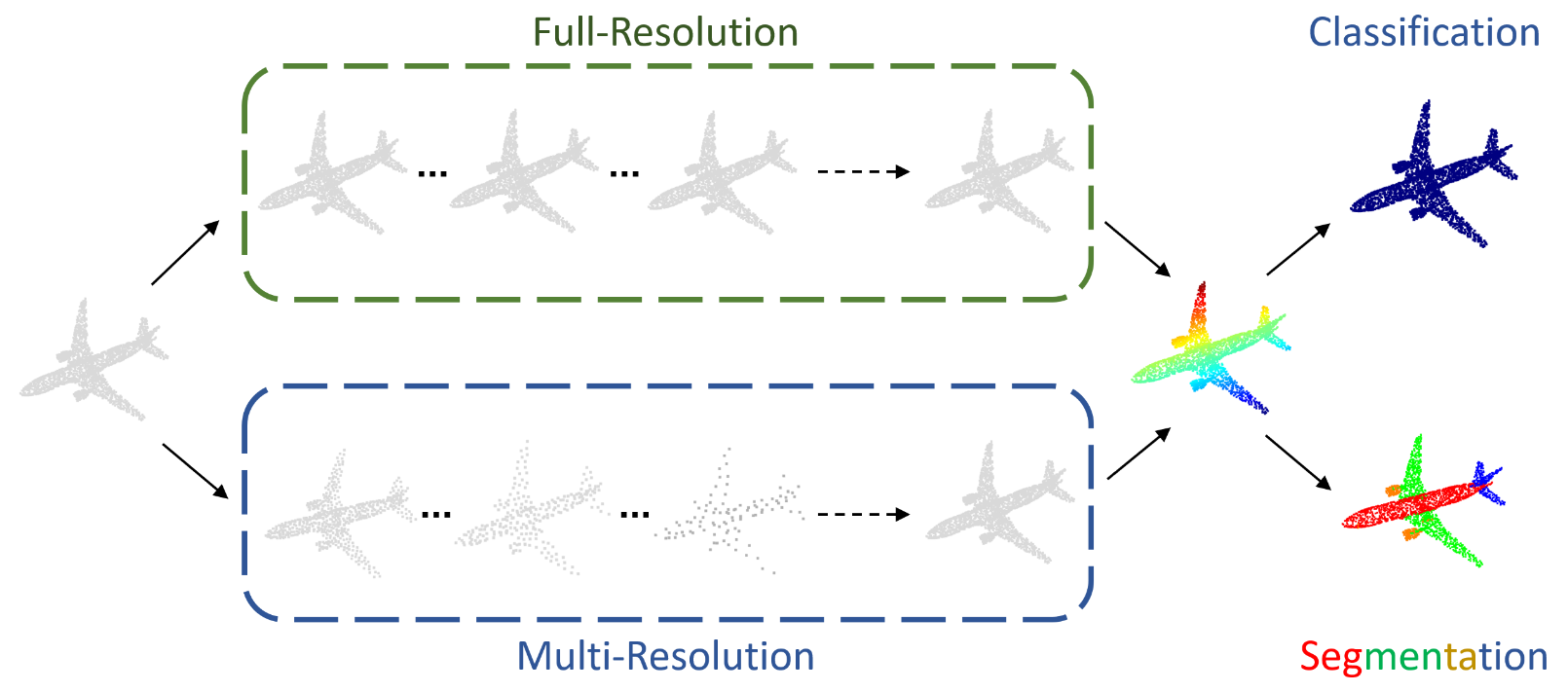}
\end{center}
\caption{A birdeyes view of our Dense-Resolution Network.}
\label{fig:intro}
\end{figure}

In order to recognize fine-grained patterns for complex objects or scenes, it is necessary to capture the local spatial context of point clouds. To represent local areas for point clouds, Qi~\etal~\cite{qi2017pointnet++} and Liu~\etal~\cite{liu2019relation} apply the Ball Query algorithm~\cite{omohundro1989five} to group local points, while Wang~\etal~\cite{wang2019dynamic} uses k-nearest neighbors (\emph{knn}) to build point neighborhoods. However, when using these methods, the performance is strongly affected by the areas of their \emph{pre-defined} neighborhoods, \ie the searching radius of a Ball Query, or the $k$ of \emph{knn}. If the area is too small, it cannot cover sufficient local patterns; if too large, the overlap may involve redundancies. DPC~\cite{engelmann2019dilated} proposes an idea of \emph{dilated point convolution} to increase the size of the receptive field without additional computational cost. Unlike previous works, we attempt to adaptively define such a local area for each point \wrt the density distribution around it, by which the point neighborhood would be more reasonable though requiring less manual intervention and parameter tuning.

\begin{figure*}
\begin{center}
\includegraphics[width=0.95\textwidth]{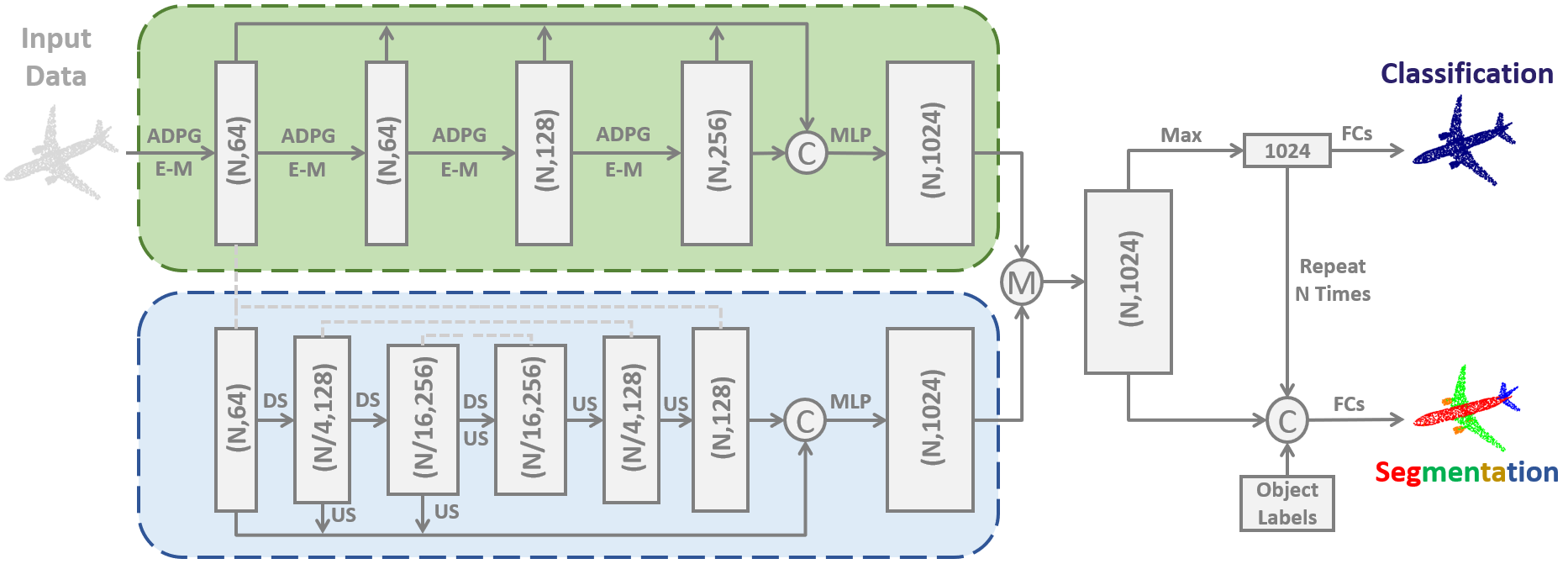}
\end{center}
\caption{Dense-resolution network architecture. For the FR branch (in green), we learn the full-resolution point cloud features through a series of Error-minimizing modules (denoted as \emph{E-M}, see Section~\ref{sec:error}) involving the Adaptive Dilated Point Grouping method (denoted as \emph{ADPG}, see Section~\ref{sec:adpg}). For the MR branch (in blue), point features of different resolutions are investigated in a down/up-sampling manner with skip connections (dotted lines). \emph{DS} and \emph{US} represent our down-sampling and up-sampling processes (more details are in Section~\ref{sec:imple} and the supplementary material), respectively. By merging the feature maps (denoted as \emph{M}, see Eq.~\ref{equ:fusion}) of the two branches, we manage point cloud classification and segmentation tasks using fully-connected (\emph{FC}) layers. \emph{C} stands for concatenating along channels.}
\label{fig:net}
\end{figure*}
Unlike 2D images whose pixels are well-organized in local neighborhoods, learning the feature representations of scattered, unordered, and irregular 3D point clouds are always challenging. Although one can construct local areas based on the spatial distances between points, the process may accumulate biases from different scales of embedding space and further affect the performance. In addition to feature encoding, an effective mechanism is also required to guide the procedure to learn local features. 

Previously, the idea of error feedback has been applied in 2D human pose estimation~\cite{carreira2016human} and image Super-Resolution (SR)~\cite{haris2018deep, liu2019hierarchical}, in order to regulate the network by compensating the estimated error. To leverage the properties of both error-feedback and CNN training mechanism, unlike the complex error-correcting structures in~\cite{Li_2019_ICCV, qiu2019geometric}, we propose an error-minimizing module with lower complexity but better performance. Meanwhile, we present a new network architecture, named Dense-Resolution Network (DRNet), for basic 3D point cloud classification and segmentation tasks. By merging feature maps of a Full-Resolution (FR) branch that investigates the full size of the point cloud and a Multi-Resolution (MR) branch that explores different resolutions of the point cloud in a novel fusion method, we can obtain more information for a comprehensive analysis. The main contributions are:

\begin{itemize}
\item We propose a novel point grouping method to find neighbors for each point adaptively, considering the density distribution of the neighbors.
\item We design an error-minimizing module leveraging the idea of error feedback mechanism to learn the local features of point clouds.
\item We introduce a new network to comprehensively represent point clouds from different resolutions.
\item We conduct thorough experiments to validate the properties and abilities of our proposals. Our results demonstrate that our approach outperforms state-of-the-art methods on three point cloud benchmarks.
\end{itemize}

\section{Related Work}
\label{sec:work}
\noindent \textbf{Local points grouping.} Contrary to the pioneer PointNet~\cite{qi2017pointnet} that relied on global features, subsequent work captured more local features in detail. PointNet++~\cite{qi2017pointnet++} firstly applied Ball Query, an algorithm for collecting possible neighbors of a particular point through a ball-like searching space centering at a point, to group local neighbors. Similarly, local features learning methods such as \cite{wang2019dynamic, engelmann2019dilated, qiu2019geometric} use another simple algorithm \emph{knn} gathering nearest neighbors based on a distance metric. 

Although Ball Query and \emph{knn} grouping are intuitive, sometimes the size of the neighborhood (\ie the receptive field of the point) is limited due to the range of searching (\ie the radius of query ball, or the value of $k$). Meanwhile, merely increasing the searching range may involve substantial computational cost. To solve this problem, DPC~\cite{engelmann2019dilated} extended regular \emph{knn} to \emph{dilated-knn}, which gathers local points over a dilated neighborhood obtained by computing the $k\cdot d$ nearest neighbors ($d$ is the dilation factor~\cite{yu2015multi}) and preserving only every $d$-th nearest point. Moreover, recent works~\cite{qi2017pointnet++, liu2019relation, yan2020pointasnl} group neighbors through query balls in different scales (\eg, multi-scale grouping) to capture information from various sizes of the local area.

However, the existing methods have some issues in common. On the one hand, the performance of grouping methods highly relies on pre-defined settings. For example, DGCNN~\cite{wang2019dynamic} provided the results under different $k$ conditions, DPC~\cite{engelmann2019dilated} compared the effects of $d$ values, and PointNet++~\cite{qi2017pointnet++} discussed the influence of the query ball radius. On the other hand, the grouping methods act on all points without considering each point or object's distinct condition. As far as we are concerned, it is necessary to find an intelligent point-level adaptive grouping method.

\noindent \textbf{Error feedback structure.} Previously in 2D computer vision, Carreira~\etal~\cite{carreira2016human} proposed a framework called Iterative Error Feedback (IEF), which minimized the error loss between current and desired outputs in the back-propagation procedure. In contrast to~\cite{carreira2016human}, the methods in~\cite{haris2018deep, liu2019hierarchical} complimented the output with a back-projection unit in the forward procedure. For 3D point clouds, PU-GAN~\cite{Li_2019_ICCV} leveraged a similar idea for point cloud generation, while ~\cite{qiu2019geometric} presented a structure with specially designed paths for prominent features learning.

Basically, current IEF structures for point clouds are redundant and implicit. Considering the complexity of 3D data, a concise and explicit IEF module is needed. More importantly, an IEF module is expected to serve two purposes in the network: first, to make the actual output approach the desired point clouds representations; second, to help the grouping process form the adaptive point neighborhoods. 

\noindent \textbf{Network architecture for point cloud learning.}   To realize different computer vision tasks using deep learning, many network architectures have been introduced: VGG~\cite{simonyan2014very}, ResNet~\cite{he2016deep}, \etc. Besides, some works tried different image resolutions for more clues; for example, the fully convolutional network~\cite{long2015fully} keeps the full size of an image, deconvolution network~\cite{noh2015learning} steps into lower resolutions, and HRNet~\cite{wang2019deep} shares the features among different resolutions.

As for 3D point clouds, two popular architectures are 1) PointNet++\cite{qi2017pointnet++}, which downsamples the point clouds using Farthest Point Sampling (FPS) and upsamples using Feature Propagation (FP), and 2) a fully convolutional network,  which learns point-wise features from multiple embedding space scales, for example, DGCNN~\cite{wang2019dynamic} dynamically updates the crafted point graph around each point. Different from the above mentioned methods, our approach exploits more clues through dense connections between various resolutions of the point clouds. Furthermore, we investigate the characteristics of multi-resolutional features, and then develop a better merging behavior for the feature maps. In general, our DRNet adaptively encodes the local context from more resolutions of point clouds, by which fine-grained output representations benefit point cloud classification and segmentation tasks.

\begin{algorithm}
\caption{The \emph{forward pass} pipeline of\newline
    \hspace{\linewidth} Adaptive Dilated Point Grouping}\label{alg:adpg}
\textbf{input:} feature map $\mathcal{P}_{{N}\times{c}} = [{p_1}^T,\;{p_2}^T,\;...\;,\;{p_N}^T ]$ in $c$-dimensional space.\\
\textbf{parameters:} the number of neighbors $k$, and an empirical maximum dilation factor $d_{max}$.\\
\textbf{output:} the matrix $\mathcal{I}_{{N}\times{k}}$, indices of the selected $k$ neighbors for the point cloud.\\
\For{each point cloud}{
    \textbf{search} for the $(k\cdot d_{max})$ candidate neighbors based on $\mathcal{P}_{{N}\times{c}}$, get the candidate metric values ${E}_{{N}\times{(k\cdot d_{max})}}$ and the indices $\mathcal{I}_{{N}\times{(k\cdot d_{max})}}$\;
    \textbf{learn} the dilation factors $\mathcal{D}_{{N}}$ based on the metrics ${E}_{{N}\times{(k\cdot d_{max})}}$, where: $d_i\in \mathbb{Z}$, $d_i\in [1,d_{max}]$, $\mathcal{D}_{{N}} = [{d_1},\;...\;,\;{d_i},\;...\;,\;{d_N}]^T$\;
    \textbf{group} the indices $\mathcal{I}_{{N}\times{k}}$ of the $k$ neighbors from $\mathcal{I}_{{N}\times{(k\cdot d_{max})}}$ based on $\mathcal{D}_{{N}}$\;
}
\end{algorithm}
\section{Approach}
CNN-based learning of 3D data has become more intuitive due to the introduction of multi-layer perceptrons (MLPs)~\cite{qi2017pointnet} that directly process point clouds. Primarily, an MLP, $\mathcal{M}(\cdot)$, is described as a composite operation of 1-by-1 convolution with a possible batch normalization~\cite{ioffe2015batch} (BN) and an activation (\eg, ReLU) on the feature map.

In addition, recent works~\cite{wang2019dynamic, engelmann2019dilated, yan2020pointasnl} craft regional patterns to record more local details via a graph around each point $p_i\in\mathbb{R}^c$, based on both the absolute position of the centroid and relative positions of the neighbors in $c$-dimensional feature space. Specifically, the crafted graph ($\mathcal{G}$) of the centroid $p_i$ is formulated as: ${\mathcal{G}}(p_i) = (p_i, p_j-p_i);$ where ${\forall}p_j\in Ni(p_i).$ Usually, the quality of the information provided by $\mathcal{G}(p_i)$ highly depends on the neighbors, ${\forall}p_j\in Ni(p_i)$, that are found by the grouping method.  Hence, we expect a better grouping method for $\mathcal{G}(p_i)$.

\subsection{Adaptive Dilated Point Grouping}
\label{sec:adpg}
The two popular grouping methods \ie Ball Query and k-nearest neighbors (\emph{knn}) (see Section~\ref{sec:intro}) have shortcomings (as analyzed in Section~\ref{sec:work}), and to overcome these issues, here we propose a novel grouping method named Adaptive Dilated Point Grouping (ADPG), which is shown in Algorithm~\ref{alg:adpg}. ADPG aims to generate the indices of neighbors $\mathcal{I}_{{N}\times{k}}$ for the points, given a feature map $\mathcal{P}_{{N}\times{c}}$ of the point cloud and consists of the following three main procedures.

\noindent \textbf{Searching.} The first step of ADPG is searching candidate neighbors for the points. In this paper, we introduce a solution capable of addressing common scales of point cloud data. We define the pairwise Euclidean distances ${E}_{{N}\times{N}}$ in feature space as our metric, which indicates the point density distribution to a certain extent. As for a $N\times c$ size feature map $\mathcal{P}$, the pairwise Euclidean distances are:
$
    {E}_{{N}\times{N}} = diag(\mathcal{P}\mathcal{P}^T)\cdot\vec{1} + {\vec{1}}^T\cdot {diag(\mathcal{P}\mathcal{P}^T)}^T - 2\mathcal{P}\mathcal{P}^T,
$
where $\vec{1}$ means a $1\times N$ row vector of ones, and $diag(\cdot)$ forms a $N\times 1$ column vector whose entries are the $N$ diagonal elements of a $N\times N$ square matrix. 

According to the calculated distances metric, we can easily identify the $k\cdot d_{max}$ candidate nearest neighbors of each point. In our implementation, we sort the rows of ${E}_{{N}\times{N}}$ in ascending order, and retain the metric values and indices of the first $k\cdot d_{max}$ elements. Therefore, the elements with the smallest $k\cdot d_{max}$ values in each row of ${E}_{{N}\times{N}}$ are identified as candidate neighbors for each point. Meanwhile, the metric values and indices of the searched candidate neighbors are recorded as ${E}_{{N}\times{(k\cdot d_{max})}}$ and $\mathcal{I}_{{N}\times{(k\cdot d_{max})}}$, respectively. Besides, our implementation is also flexible; that is, the choices for metrics (\eg, density or geometric similarities) and searching techniques (\eg, FLANN~\cite{muja2009fast} for the sake of efficiency in large-scale point cloud data) can be easily integrated as needed.

\noindent \textbf{Learning.} In order to construct a dilated neighborhood for each point adaptively, it is necessary to determine a dilation factor~\cite{yu2015multi} for each point based on known information of its candidate neighbors. In practice, we learn the dilation factors based on ${E}_{{N}\times{(k\cdot d_{max})}}$ and CNN-related operations.  

To be specific, we apply an MLP ($\mathcal{M}$) and a sigmoid function ($\sigma$) to the metric values of candidates ${E}_{{N}\times{(k\cdot d_{max})}}$, in order to summarize the information of the point distribution of local areas. Then, a projection function $\mathcal{J}$ (\eg, \emph{linear function}) can map the values to the expected numerical scale. Finally, we take a scale function $\mathcal{S}$ (\eg, \emph{round} to assign a dilation factor, $\mathcal{D}_{{N}}$\footnote{More implementing details are in the supplementary material.}, for each point according to the summarized information:
\begin{equation}
    \mathcal{D}_{{N}} = \mathcal{S}\bigg(\mathcal{J}\Big(\sigma\big(\mathcal{M}({E}_{{N}\times{(k\cdot d_{max})}})\big)\Big)\bigg).
\end{equation}

\noindent \textbf{Grouping.} As each point has a corresponding dilation factor, we pick up every $d_i$-th index of candidate indices $\mathcal{I}_{{N}\times{(k\cdot d_{max})}}$ to form the selected $k$ neighbors for each point. Following a behavior similar to~\cite{engelmann2019dilated}, we obtain the final indices of local point groups $\mathcal{I}_{{N}\times{k}}.$

\subsection{Error-minimizing Module}
\label{sec:error}
Following the ADPG method, each point gathers a group of neighbors with a larger receptive field. As stated, we apply the crafted graph $\mathcal{G}$, \ie the absolute position of a centroid and relative positions of the neighbors, to encode the high-dimensional features over each neighborhood. Further projected by an MLP (with $c^\prime$ filters), the information of a local graph centering at $p_i$, is represented as:
\begin{equation}
\label{equ:graph}
    f_{{\mathcal{G}}_i} = \mathcal{M}\big(\mathcal{G}(p_i)\big) = \mathcal{M}\big((p_i, p_j-p_i)\big),
\end{equation}
where ${\forall}p_j\in {ADPG}(p_i)$ and $f_{{\mathcal{G}}_i}\in\mathbb{R}^{{c^\prime}\times k}.$

Usually, a max-pooling function is applied over the $k$ neighbors of each crafted local graph to aggregate the local context as the centroid's feature representation. However, possible bias exists in process: on the one hand, the local graphs lack geometric regularization from the initial 3D space; and on the other hand, the max-pooled features only retain prominent outlines while discarding local details in embedding space. In this case, the Iterative Error Feedback (IEF) mechanisms idea helps avoid bias accumulation during the high-dimensional feature learning process.

Let us assume that the local graph $f_{{\mathcal{G}}_i}$ indeed embeds the full information about the neighborhood, it would be possible to restore the input $p_i$ through a back-projection process $\mathcal{B}(\cdot)$. Practically, we realize the $\mathcal{B}(\cdot)$ operation through a shared 1-by-$k$ convolution followed by BN and ReLU, over the local graphs. Intuitively, this operation acts to aggregate the nodes based on learned weights of the edges in the graph, which implicitly simulates a reverse process of crafting the graph. Therefore, the back-projected feature $f_{{\mathcal{B}}_i}$ is formulated as:
$
    f_{{\mathcal{B}}_i} = \mathcal{B}(f_{{\mathcal{G}}_i}); 
$
where $f_{{\mathcal{B}}_i}\in\mathbb{R}^{c}.$

Consequently, an error feature $f_{{\mathcal{E}}_i}$ is defined as the difference between the original input $p_i$ and back-projected feature $f_{{\mathcal{B}}_i}$. In contrast to the methods in~\cite{Li_2019_ICCV, qiu2019geometric, haris2018deep, liu2019hierarchical} that correct the error by extra computations in the forward pass, we use additional $\ell_2$ loss to minimize the error, $f_{{\mathcal{E}}_i} = f_{{\mathcal{B}}_i} - p_i$, during the back-propagation pass:
\begin{equation}
\label{equ:loss}
    {\mathcal{L}}_{er} = {||f_{{\mathcal{E}}_i}||}_2.
\end{equation}

The loss in Equation~\ref{equ:loss} can constrain the feature learning during training by forcing the back-projected feature $f_{{\mathcal{E}}_i}$ to approach the original input $p_i$ inside of the module. Following such a regularization, the error and bias in the output representations can be alleviated, especially during the early stages of training. Meanwhile, compared with the regular cross-entropy loss for the whole network, each error-minimizing module's loss can provide more clues for the ADPG in corresponding feature space.

\subsection{Dense-Resolution Network Architecture}
\label{sec:dr}
Although the ADPG method and the error-minimizing module seem promising for local feature learning of point clouds, we still need a robust network architecture to leverage the potential offered by both. The architecture of our network is presented in Figure~\ref{fig:net}.

\noindent \textbf{Full-resolution branch.} We adopt the idea of basic fully convolutional architecture as the full-resolution (FR) branch of our network. The benefits can be retained based on two aspects; 1) there remains a consistent number of points in different scales of embedding space during feature learning progress; 2) it retains the per-point feature without any confusion caused by the numerical approximation in upsampling. Therefore, we expect this structure to learn comprehensive representations for point-wise features.

Specifically, the FR branch consists of the proposed error-minimizing modules in a cascaded form, which progressively learn the feature representation of each point from its adaptive neighborhood formed by ADPG in different scales of embedding space. In order to acquire a global knowledge about the abstract embedding space, the learned features from different scales are concatenated and aligned to form the output $\mathcal{F}_{FR}$ of the FR branch.    

\noindent \textbf{Multi-resolution branch.} Meanwhile, there is a limitation of $\mathcal{F}_{FR}$: it lacks channel-wise clues about semantic/shape-related information since the FR branch mainly focuses on point-wise context. To overcome this issue, we capture additional features from more resolutions of point clouds. Therefore, we propose the multi-resolution (MR) branch, a light-weight down/up-sampling structure, to investigate the lower resolutions of point clouds. Contrary to competing methods, the propagated features and skip links are densely connected to enhance the relations between multiple point cloud resolutions and feature embedding scales. The output $\mathcal{F}_{MR}$ of the MR branch captures thorough channel-wise information about the point clouds.

\noindent \textbf{Features merging.} To leverage the information gathered from both FR and MR branches, it is necessary to find a reasonable merging technique for the two feature maps, \ie $\mathcal{F}_{FR}$ and $\mathcal{F}_{MR}$. Usually, CNNs combine the feature maps by concatenation, summation, or multiplication. These regular operations treat the feature maps equally, without taking their properties into account. Instead, we prefer merging the FR and MR outputs in a unique manner.

Given the advantages of FR and MR branches that we analyzed before, $\mathcal{F}_{FR}$ is applied as the basis of per-point feature representation. In addition, the channel-wise information of $\mathcal{F}_{MR}$ is derived to enhance $\mathcal{F}_{FR}$. Empirically, we use a max-pooling and an MLP to summarize the knowledge of $\mathcal{F}_{MR}$ channels. After a sigmoid activation $\sigma$, the channel-wise enhancement on the per-point context of $\mathcal{F}_{FR}$ can be realized by multiplication. The final output of our dense-resolution (DR) network follows:
\begin{equation}
\label{equ:fusion}
    \mathcal{F}_{DR}\; = \;\mathcal{F}_{FR}\; \times \; \sigma\Big(\mathcal{M}\big({\max_{N}(\mathcal{F}_{MR})}\big)\Big).
\end{equation}

\noindent \textbf{Loss function.} Based on the output feature map ($\mathcal{F}_{DR}$), the fully-connected (FC) layers regress the confidence scores for all possible categories. In addition to the basic cross-entropy loss ($\mathcal{L}_{ce}$), the weighted losses of the error-minimizing modules are incorporated. For the DRNet with $M$ error-minimizing modules in its FR branch, by applying Equation~\ref{equ:loss} and the hyper-parameter $w_i$ as weight, the overall loss is formulated as:
\begin{equation}
\label{equ:overall_loss}
    \mathcal{L} = \mathcal{L}_{ce} + \sum_{i=1}^{M} w_i\cdot \mathcal{L}_{{er}_{i}}.
\end{equation}
\section{Experiments}
In this section, our implementation details are provided, including network parameters, training settings, datasets, \etc. By comparing the experimental results with other state-of-the-art methods, we analyze performance quantitatively. Further, we present ablation studies and visualizations to illustrate the properties of our approach.

\subsection{Implementation}
\label{sec:imple}
\noindent \textbf{Network details.}  The FR branch of our DRNet is a series of error-minimizing modules extracting features at different scales of embedding space: \ie 64, 128, and 256, as in Figure~\ref{fig:net}. Empirically, we adopt $k=20$ and $d_{max}=5$ as in~\cite{wang2019dynamic, engelmann2019dilated}. The FR output $\mathcal{F}_{FR}$ is an MLP projected concatenation of the modules' outputs. For the MR branch, we apply the widely-used farthest point sampling (FPS) and feature propagation (FP)~\cite{qi2017pointnet++, liu2019relation, Liu_2019_ICCV} for downsampling and upsampling, respectively. Further, single-layer MLPs are used for channel alignment together with the mentioned operations. The MR branch starts from the first output of FR in N size; after that, two lower resolutions: N/4 and N/16, are investigated through the regular \emph{knn} and local graph encoding as Equation~\ref{equ:graph}. Different from other approaches, more propagated features and dense skip connections are employed to enhance the relations between different point resolutions and feature spaces. Compared with the FR, the MR branch\footnote{More information about the implementation is provided in the supplementary material. \label{footnote}} is light-weight due to the fewer scales of embedding space, the limited number of points, and the operations with fewer learnable weights.

The output $\mathcal{F}_{DR}$ is obtained by following Equation~\ref{equ:fusion}. For the classification task, we apply a max-pooling function and Fully Connected (FC) layers to regress confidence scores for all possible categories. In terms of the segmentation task, we attach the max-pooled feature to each point feature of $\mathcal{F}_{DR}$ and further predict each point's semantic label with FC layers being applied. 

For the loss function, empirically, a larger weight is set for the first error-minimizing module, \ie $w_1$, since its output affects both branches and constrains the network learning initially. In contrast, the weights for other modules can be smaller since they are less critical. Although the additional loss is involved, cross-entropy loss still contributes the most to the training\footref{footnote}. We implement the project with PyTorch and Python; all experiments are conducted on Linux and GeForce RTX 2080Ti GPUs.\footnote{The code and models are available at \url{https://github.com/ShiQiu0419/DRNet}}

\begin{table*}
\begin{center}
\resizebox{0.9\textwidth}{!}{
\begin{tabular}{|c| c| c c c c c c c c c c c c c c c c|}
\hline
&\textbf{overall}  & air & bag & cap  & car  & chair  & ear & guitar & knife & lamp & laptop & moto & mug & pistol  & rocket & skate & table  \\
&\textbf{mIoU}  & plane &  &   &   &   & phone &  &  &  &  & bike &  &   &  & board &   \\\hline\hline
\# shapes  & 16881 & 2690   & 76 & 55  & 898  & 3758  & 69    & 787  & 392  & 1547     & 451  & 202  & 184   & 283  & 66    & 152  & 5271     \\ \hline\hline
PointNet \cite{qi2017pointnet} & 83.7 &83.4 &78.7 &82.5 &74.9 &89.6 &73.0 &91.5 &85.9 &80.8 &95.3 &65.2 &93.0 &81.2 &57.9 &72.8 &80.6     \\ 
A-SCN \cite{xie2018attentional} & 84.6&83.8 &80.8 &83.5 &79.3 &90.5 &69.8 &91.7 &86.5 &82.9 &\textbf{96.0} &69.2 &93.8 &82.5 &62.9 &74.4 &80.8     \\ 
SO-Net \cite{li2018so} &84.6&81.9 &83.5 &84.8 &78.1 &90.8 &72.2 &90.1 &83.6 &82.3 &95.2 &69.3 &94.2 &80.0 &51.6 &72.1 &82.6 \\ 
PointNet++ \cite{qi2017pointnet++}   & 85.1 & 82.4  & 79.0  & 87.7  & 77.3  & 90.8  & 71.8  & 91.0  & 85.9  & 83.7  & 95.3  & 71.6  & 94.1  & 81.3  & 58.7  & 76.4  & 82.6   \\ 
PCNN \cite{atzmon2018point} &85.1&82.4 &80.1 &85.5 &79.5 &90.8 &73.2 &91.3 &86.0 &85.0 &95.7 &73.2 &94.8 &83.3 &51.0 &75.0 &81.8 \\ 
DGCNN \cite{wang2019dynamic}  & 85.2 & 84.0   & 83.4 & 86.7  & 77.8  & 90.6  & 74.7    & 91.2  & 87.5  & 82.8     & 95.7  & 66.3  & 94.9   & 81.1  & \textbf{63.5}    & 74.5  & 82.6     \\ 
P2Sequence \cite{liu2019point2sequence} &85.2 &82.6 &81.8 &87.5 &77.3 &90.8 &77.1 &91.1 &86.9 &83.9 &95.7 &70.8 &94.6 &79.3 &58.1 &75.2 &82.8 \\
SpiderCNN \cite{xu2018spidercnn} &85.3 &83.5 &81.0 &87.2 &77.5 &90.7 &76.8 &91.1 &87.3 &83.3 &95.8 &70.2 &93.5 &82.7 &59.7 &75.8 &82.8 \\ 
PointASNL \cite{yan2020pointasnl} &86.1 &84.1 &84.7 &87.9 &\textbf{79.7} &\textbf{92.2} &73.7 &91.0 &87.2 &84.2 &95.8 &\textbf{74.4} &\textbf{95.2} &81.0 &63.0 &76.3 &83.2 \\ 
RS-CNN \cite{liu2019relation} &86.2 &83.5 &84.8 &\textbf{88.8} &79.6 &91.2 &\textbf{81.1} &91.6 &88.4 &\textbf{86.0} &\textbf{96.0} &73.7 &94.1 &\textbf{83.4} &60.5 &\textbf{77.7} &83.6  \\
\hline\hline
\textbf{Ours}  &\textbf{86.4} &\textbf{84.3} &\textbf{85.0} &88.3 &79.5 &91.2 &79.3 &\textbf{91.8} &\textbf{89.0} &85.2 &95.7 &72.2 &94.2 &82.0 &60.6 &76.8 &\textbf{84.2}   \\ \hline
\end{tabular}
}
\end{center}
\caption{Part segmentation results (mIoU(\%)) on the \emph{ShapeNet Part} dataset. }
\label{tab:seg}
\end{table*}
\noindent \textbf{Training strategy.} For classification, Stochastic Gradient Descent (SGD)~\cite{loshchilov2016sgdr} with a momentum of 0.9 is adopted as the optimizer. The learning rate decreases from 0.1 to 0.001 by cosine annealing \cite{loshchilov2016sgdr} during the 300 epochs. For segmentation, we exploit Adam~\cite{kingma2014adam} optimization for 200 epochs of training. The learning rate begins at 0.001 and gradually decays with a rate of 0.5 after every 20 epochs. The batch size for both tasks is 32. Besides, training data is augmented with random scaling and translation; the overall loss follows Equation~\ref{equ:overall_loss}. Part segmentation is evaluated with a ten-votes strategy used by recent approaches~\cite{qi2017pointnet, qi2017pointnet++, liu2019relation}.

\begin{figure}
\begin{center}
\includegraphics[width=0.95\columnwidth]{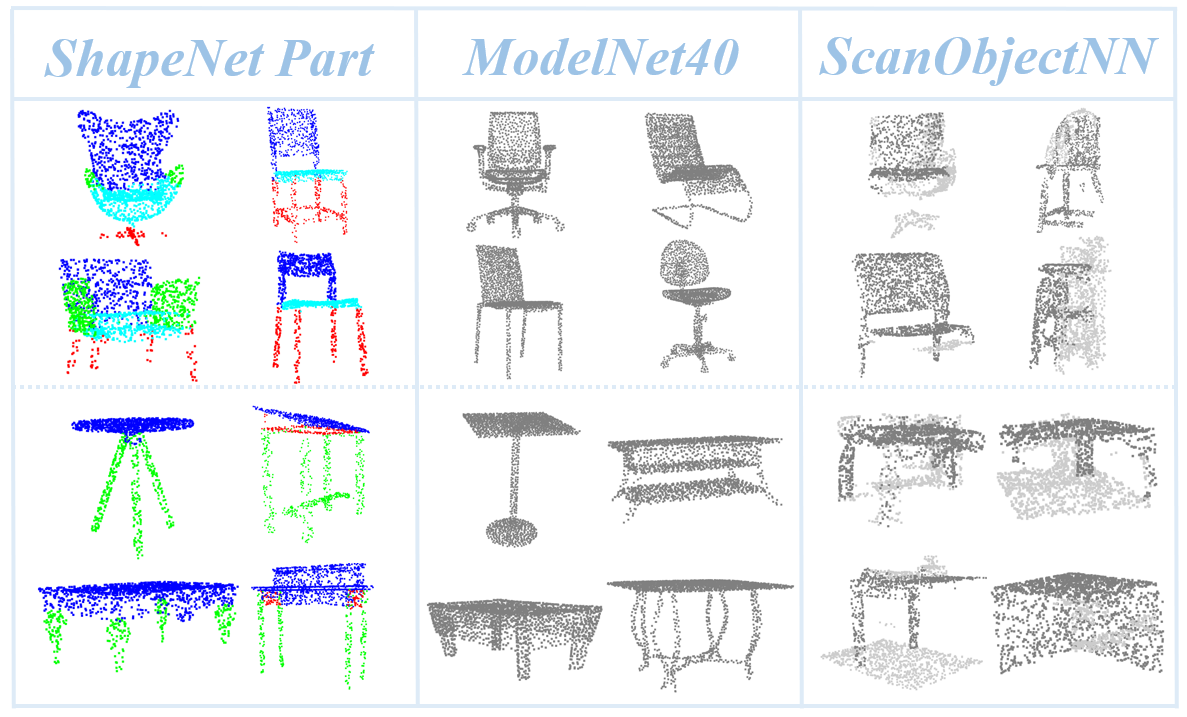}
\end{center}
\caption{Examples from the experimental datasets. The upper row shows the point clouds labeled as \emph{Chair} from the three datasets, while the lower row presents \emph{Table}. Particularly, \emph{ScanObjectNN} dataset contains background points, which are in a lighter color.} 
\label{fig:data}
\end{figure}

\noindent \textbf{Datasets.}  We test our approach on two main tasks: point cloud segmentation and classification. The ShapeNet Part dataset~\cite{yi2016scalable} is used to predict the semantic class (\emph{part label}) for each point of the object. In addition, the synthetic ModelNet40~\cite{wu20153d} dataset and the real-world ScanObjectNN \cite{Uy_2019_ICCV} dataset are used to identify the category of the object. Figure~\ref{fig:data} presents some examples from the datasets.
\begin{itemize}
 \item \textbf{ShapeNet Part.} 
 The dataset has 16,881 object point clouds in 16 categories, where each point is labeled as one of 50 parts. As the primary dataset for our experiments, we follow the official data split~\cite{chang2015shapenet}. We input the 3D coordinates of 2048 points for each point cloud and feed the object label before FC layers during training. In terms of the metric for evaluation, we adopt Intersection-over-Union (\ie IoU). The IoU of the shape is calculated by the mean value of IoUs of all parts in that shape, and mIoU (\ie mean IoU) is the average of IoUs for all testing shapes.
 
 \item \textbf{ModelNet40.}  
 It is a popular dataset because of regular and clean shapes. There are 12,311 meshes in 40 classes, with 9,843 for training and 2,468 for testing. Corresponding point clouds are generated by uniformly sampling from the surfaces, translating to the origin, and scaling within a unit sphere~\cite{qi2017pointnet}. In our case, only the 3D coordinates of 1024 points for each point cloud have been used.
 
 \item \textbf{ScanObjectNN.} This real-world object dataset is recently published. Although it has over 15,000 objects in only 15 categories, it is practically more challenging due to the background complexity, object partiality, and different deformation variants. We conduct the experiment using its most challenging variant, \emph{PB\_T50\_RS}, with background points.
\end{itemize}

\begin{table}
\begin{center}
\resizebox{0.8\columnwidth}{!}{
\begin{tabular}{|l|c|c|c|}
\hline
method             & input type  & \#points & accuracy \\ \hline\hline
PointNet \cite{qi2017pointnet}           & coords      & $1k$       & 89.2                   \\ 
A-SCN \cite{xie2018attentional}                & coords      & $1k$       & 90.0                  \\        
PointNet++ \cite{qi2017pointnet++}         & coords & $1k$       & 90.7                       \\
SO-Net \cite{li2018so}             & coords      & $2k$       & 90.9                   \\ 
PointCNN \cite{li2018pointcnn}           & coords      & $1k$       & 92.2                   \\
PCNN \cite{atzmon2018point}               & coords      & $1k$       & 92.3                       \\
SpiderCNN \cite{xu2018spidercnn}         & coords+norms      & $1k$       & 92.4                      \\
PointConv \cite{wu2019pointconv}                           & coords+norms      & $1k$       & 92.4                 \\
P2Sequence \cite{liu2019point2sequence}               & coords      & $1k$       & 92.6         \\
DensePoint \cite{Liu_2019_ICCV}         & coords      & $1k$       & 92.8            \\
RS-CNN \cite{liu2019relation}           & coords      & $1k$       & 92.9                       \\ 
DGCNN \cite{wang2019dynamic}              & coords      & $1k$       & 92.9                    \\ 
KP-Conv \cite{Thomas_2019_ICCV}              & coords      & $7k$       & 92.9                    \\
PointASNL \cite{yan2020pointasnl}              & coords      & $1k$       & 92.9                   \\
 \hline\hline
\textbf{Ours}   & coords&$1k$&\textbf{93.1}  \\ \hline

\end{tabular}
}
\end{center}
\caption{Overall classification accuracy (\%) on \emph{ModelNet40} dataset. ($coords$: 3D coordinates, $norms$: surface normal vectors of the points, $k$:$\times 2^{10}$)}
\label{tab:cls}
\end{table}

\begin{figure*}
\begin{center}
\includegraphics[width=0.8\textwidth]{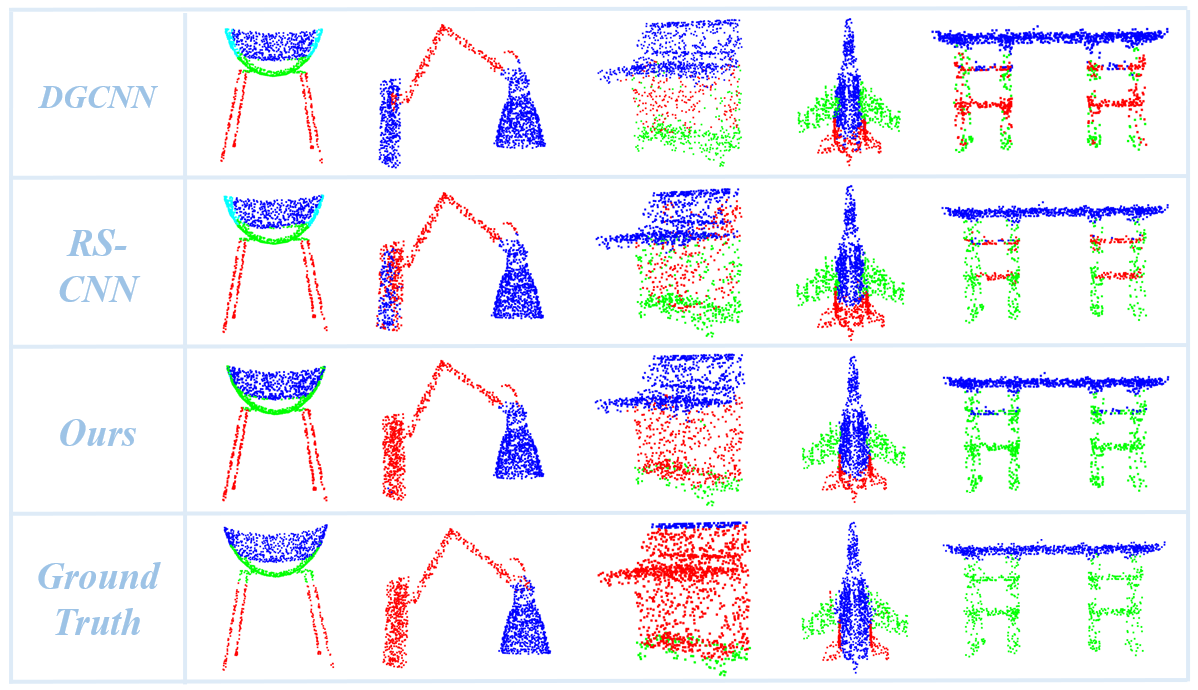}
\end{center}
\caption{Examples of the part segmentation results. (DGCNN:~\cite{wang2019dynamic}, RS-CNN:~\cite{liu2019relation})}
\label{fig:seg_vis}
\end{figure*}

\begin{table*}
\begin{center}
\resizebox{0.9\textwidth}{!}{
\begin{tabular}{|c|c|c|c c c c c c c c c c c c c c c|}
\hline
& overall acc. & avg class acc. & bag  & bin  & box  & cabinet & chair & desk & display & door & shelf & table & bed  & pillow & sink & sofa & toilet \\ \hline\hline
\# shapes  &      -       &      -         & 298  & 794  & 406  & 1344    & 1585  & 592  & 678     & 892  & 1084  & 922   & 564  & 405    & 469  & 1058 & 325    \\ \hline\hline
3DmFV \cite{ben20183dmfv}      & 63.0           & 58.1           & 39.8 & 62.8 & 15.0 & 65.1    & 84.4  & 36.0 & 62.3    & 85.2 & 60.6  & 66.7  & 51.8 & 61.9   & 46.7 & 72.4 & 61.2   \\ \hline
PointNet \cite{qi2017pointnet}   & 68.2         & 63.4           & 36.1 & 69.8 & 10.5 & 62.6    & 89.0  & 50.0 & 73.0    & \textbf{93.8} & 72.6  & 67.8  & 61.8 & 67.6   & 64.2 & 76.7 & 55.3   \\ \hline
SpiderCNN \cite{xu2018spidercnn}  & 73.7         & 69.8           & 43.4 & 75.9 & 12.8 & 74.2    & 89.0  & 65.3 & 74.5    & 91.4 & 78.0  & 65.9  & 69.1 & 80.0   & 65.8 & 90.5 & 70.6   \\ \hline
PointNet++ \cite{qi2017pointnet++} & 77.9         & 75.4           & 49.4 & \textbf{84.4} & 31.6 & 77.4    & 91.3  & \textbf{74.0} & 79.4    & 85.2 & 72.6  & 72.6  & 75.5 & \textbf{81.0}   & \textbf{80.8} & 90.5 & \textbf{85.9}   \\ \hline
DGCNN \cite{wang2019dynamic}      & 78.1         & 73.6           & 49.4 & 82.4 & 33.1 & \textbf{83.9}    & 91.8  & 63.3 & 77.0    & 89.0 & 79.3  & \textbf{77.4}  & 64.5 & 77.1   & 75.0 & 91.4 & 69.4   \\ \hline
PointCNN \cite{li2018pointcnn}   & 78.5         & 75.1           & 57.8 & 82.9 & 33.1 & 83.6    & \textbf{92.6}  & 65.3 & 78.4    & 84.8 & \textbf{84.2}  & 67.4  & \textbf{80.0} & 80.0   & 72.5 & \textbf{91.9} & 71.8   \\ \hline\hline
\textbf{Ours}       & \textbf{80.3}         & \textbf{78.0}           & \textbf{66.3} & 81.9 & \textbf{49.6} & 76.3    & 91.0  & 65.3 & \textbf{92.2}    & 91.4 & 83.8  & 71.5  & 79.1 & 75.2   & 75.8 & \textbf{91.9} & 78.8   \\ \hline
\end{tabular}
}
\end{center}
\caption{Classification results (\%) on \emph{ScanObjectNN} dataset. }
\label{tab:scanobjectnn}
\end{table*}
\subsection{Results}
\noindent \textbf{Segmentation.} Table~\ref{tab:seg} shows the results of related works reported in overall mIoU, which is the most critical evaluation metric on the ShapeNet Part dataset. On the whole, our network achieves 86.4\% and outperforms other state-of-the-art algorithms based on similar experimental settings. For evaluations inside each class, we surpass others in five out of 16 categories. Especially in categories with a large number of samples, \eg, airplane, chair, or table, we perform even better (two out of these three classes) than others. In Figure~\ref{fig:seg_vis}, we provide some samples of our part segmentation results comparing with DGCNN~\cite{wang2019dynamic} and RS-CNN~\cite{liu2019relation}.

\noindent \textbf{Classification.}   Table~\ref{tab:cls} presents the overall accuracy of the classification on the synthetic object dataset: ModelNet40. Specifically, we achieve 93.1\% for overall classification accuracy and exceed other state-of-the-art results with similar input. Essentially, our method performs better than others using more input points or features. 

\begin{figure*}
\begin{center}
\includegraphics[width=0.9\textwidth]{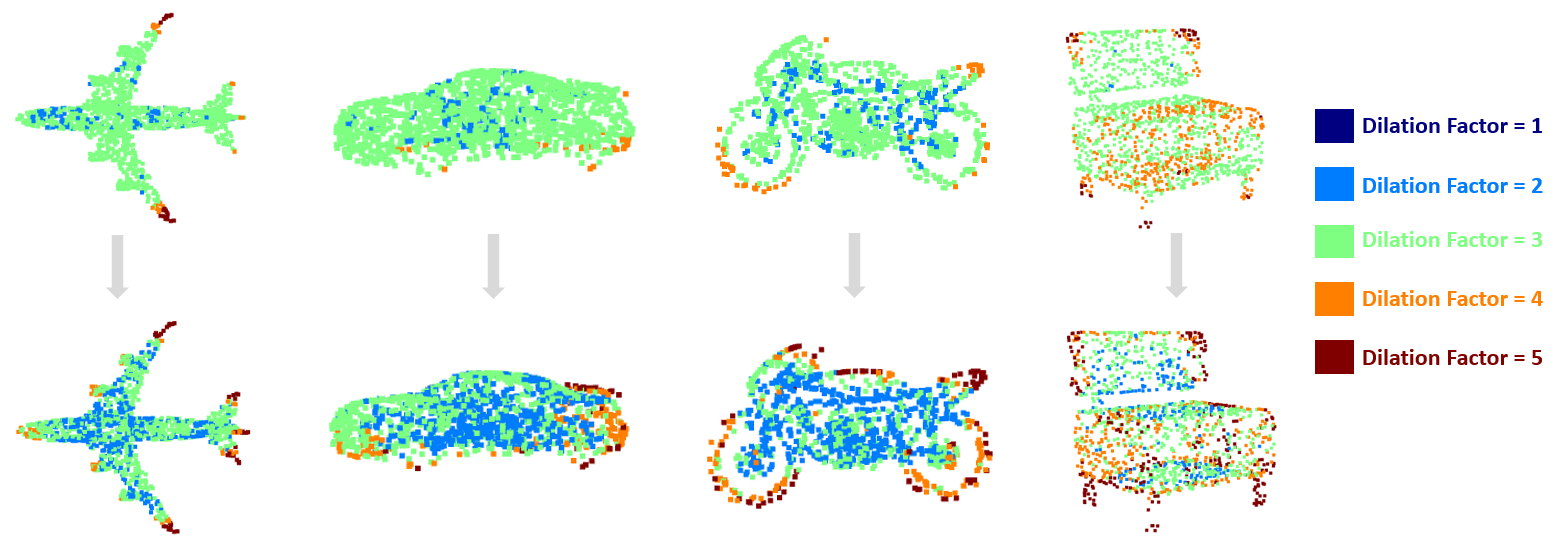}
\end{center}
\caption{Learned dilation factors by the ADPG method. For each point cloud, ADPG assigns larger dilation factors for the points in sparse areas. As the network goes deeper, ADPG regulates the dilation factors of the points. (First-row: the learned dilation factors in a shallow layer of our network. Second-row: in a deep layer.)}
\label{fig:visual}
\end{figure*}

Table~\ref{tab:scanobjectnn} presents our results on the ScanObjectNN dataset, which contains practical scans of real-world objects as Figure~\ref{fig:data} indicates. To be concrete, both overall accuracy 80.3\% and average class accuracy 78.0\% of our approach are significantly higher than all results on its official leaderboard~\cite{scanWeb}. Typically, we lead in four (\emph{bag, box, display, and sofa}) out of the 15 categories. The inference time of our basic classification model running on a single GeForce RTX 2080Ti GPU is about \emph{19.2ms}.

\subsection{Ablation Studies}
\label{sec:abl}
\noindent \textbf{Visualization of learned dilation factors.}   
Figure~\ref{fig:visual} illustrates the effects of our ADPG method, where the color of the points corresponds to the learned dilation factor. Intuitively, the advantages of ADPG can be observed from two aspects: Firstly, for each point cloud, ADPG tends to assign larger dilation factors to points that have relatively sparse local point distributions (\eg, on corners/boundaries/edges) because they need larger neighborhoods for more comprehensive local feature learning. Secondly, within the cascaded structure, ADPG regulates the points' dilation factors in deep layers and turns out to have smaller dilation factors in dense local distribution (\eg, on flat surfaces/central areas), most probably to constrain the neighborhoods and avoid outliers. Unlike regular \emph{knn}/Ball Query/\emph{dilated-knn}, which defines a limited and fixed neighborhood for all points in all layers, our ADPG works adaptively and reasonably as expected.

\noindent \textbf{Effects of components.}     We conduct an ablation study about the effects of the network components: the architecture, grouping method, and the error-minimizing module. We run tests on the ShapeNet Part dataset under the same settings, and Table~\ref{tab:modules} presents the results. Comparing model 1\&2 to model 0 and model 4 to model 3, we observe that the error-minimizing module with ADPG applied can significantly improve the part segmentation's network performance. Although the multi-resolution branch is not able to learn the features as comprehensively as a full-resolution branch does, we can take advantage of both by combining them as a dense-resolution network (model 5).

\begin{table}
\begin{center}
\resizebox{0.9\columnwidth}{!}{
\begin{tabular}{c||c|c|c||c}
\Xhline{3\arrayrulewidth}
model    & Network & ADPG & E-M module & overall mIoU \\\hline
0        & $FR$     & -    & -     & 85.2         \\
1        & $FR$ & -  &\checkmark          & 85.4      \\
2        & $FR$ & \checkmark & \checkmark      & 85.6        \\
3        & $MR$ & - & -      & 84.9         \\
4        & $MR$ & \checkmark & \checkmark      & 85.3         \\
\textbf{5}        & $\emph{\textbf{DR}}$      & \checkmark    & \checkmark     & \textbf{86.0}     
  \\\Xhline{3\arrayrulewidth}      
\end{tabular}
}
\end{center}
\caption{Ablation study about the effects of different network components on ShapeNet Part (\%). ($FR$: Full-Resolution branch only, $MR$: Multi-Resolution branch only, $DR$: Dense-Resolution Network, ADPG: Adaptive Dilated Point Grouping method, E-M module: Error-minimizing module for local points.)}
\label{tab:modules}
\end{table}
\noindent \textbf{Merging the feature maps.}   Both FR and MR have properties as mentioned, so we need to find an effective way to unify the benefits. We test simple ways of merging the features of $\mathcal{F}_{FR}$ and $\mathcal{F}_{MR}$, \ie concatenating them in channel-wise, adding and multiplying them in element-wise. Comparing the results of model 2\&3\&4 to model 0 in Table~\ref{tab:output}, we observe that the simple ways of merging may not improve performance. In contrast, channel-wise enhancing of the $\mathcal{F}_{FR}$ using $\mathcal{F}_{MR}$ (model 5) can improve a bit because of the reasons explained in Section~\ref{sec:dr}. With ten-votes testing, the overall mIoU can boost to 86.4\%. 
\begin{table}
\begin{center}
\resizebox{0.9\columnwidth}{!}{
\begin{tabular}{c||c|c||c}
\Xhline{3\arrayrulewidth}
model    & Network & $\mathcal{F}_{mer}$ & overall mIoU \\\hline
0        & $FR$     & $\mathcal{F}_{FR}$       & 85.6         \\
1        & $MR$ & $\mathcal{F}_{MR}$            & 85.3      \\
2        & $DR$ & $Concat(\mathcal{F}_{FR}, \mathcal{F}_{MR})$       & 85.7        \\
3        & $DR$ & $\mathcal{F}_{FR}+\mathcal{F}_{MR}$       & 85.6         \\
4        & $DR$      & $\mathcal{F}_{FR}\odot\mathcal{F}_{MR}$      & 85.6\\
\textbf{5}        & $\emph{\textbf{DR}}$      & $\mathcal{F}_{DR}$      & \textbf{86.0}
  \\\Xhline{3\arrayrulewidth}      
\end{tabular}
}
\end{center}
\caption{Ablation study about the different forms of merged feature $\mathcal{F}_{mer}$ on ShapeNet Part (\%). ($FR$: Full-Resolution branch only, $MR$: Multi-Resolution branch only, $DR$: Dense-Resolution Network, $\mathcal{F}_{FR}$: the output of FR, $\mathcal{F}_{MR}$: the output of MR, $\odot$: element-wise multiplication, $\mathcal{F}_{DR}$: merging as in~Equation~\ref{equ:fusion}.)}
\label{tab:output}
\end{table}
\section{Conclusion} In this work, we propose a Dense-Resolution Network for point cloud analysis, which leverages information from different resolutions of the point clouds. Specifically, the Adaptive Dilated Point Grouping method is introduced to realize a flexible point grouping based on the density distribution. Moreover, an error-minimizing module and corresponding loss are presented to capture local information and guide the training network. We conduct experiments and provide ablation studies on both point cloud segmentation and classification benchmarks. The experimental results outperform competing state-of-the-art methods on ShapeNet Part, ModelNet40, and ScanObjectNN datasets. The quantitative reports and qualitative visualizations demonstrate the advantages of our approach.
{\small
\bibliographystyle{ieee_fullname}
\bibliography{egbib}
}
\clearpage
\appendix
\noindent\textbf{\Large{Supplementary Material}}
\section{Overview}
In this supplementary material, we present more contents of our paper \emph{Dense-Resolution Network for Point Cloud Classification and Segmentation}. To be specific, we provide the implementation of the \emph{Adaptive Dilated Point Grouping (ADPG)} method and \emph{loss function} for the experiments. Besides, we show the details of our Multi-resolution (MR) branch. By comparing the relevant model parameters with others on \emph{ModelNet40} dataset, we discuss the complexity of our network.
\section{Implementation}
In the main paper, we introduce the pipeline for the \emph{ADPG} method and the design of \emph{loss function} for training. In this section, we provide more practical details in our experiments.
\subsection{ADPG Learning Process}
In practice, $d_{max}$ is an empirical parameter which may vary between the data scales or networks. In our experiments, we set $d_{max}=5$ for ShapeNet Part, ModelNet40 and ScanObjectNN datasets, since they share the similar scales of point clouds. 

Assume that we already have the indices $\mathcal{I}_{{N}\times{(k\cdot d_{max})}}$ and metrics ${E}_{{N}\times{(k\cdot d_{max})}}$ for $k\cdot d_{max}$ candidates, the crucial step of \emph{ADPG} is to learn a certain dilation factor for each point based on the known information. In Section 3.1 of the paper, we present the general description for this process:
$$
    \mathcal{D}_{{N}} = \mathcal{S}\bigg(\mathcal{J}\Big(\sigma\big(\mathcal{M}({E}_{{N}\times{(k\cdot d_{max})}})\big)\Big)\bigg)
$$

Specifically, we apply a two-layer Multi-Layer-Perceptron (MLP $\mathcal{M}$): $Conv_{[1,1]}^{\big\{({k\cdot d_{max}}/2),1\big\}}$ first, then activate corresponding \emph{negative} values using a \emph{logistic function}: $y = 1/(1+e^{-x})$. Since the values are in between 0 and 1, $\mathcal{J}$ can further enlarge the variance by projecting them to another interval. Here we expect the values to be in $[0.5,5.5]$, thus a simple linear projection function $y = 5\cdot x + 0.5$ serves as $\mathcal{J}$. Finally, we adopt \emph{round function} as $\mathcal{S}$ to scale the continuous values in $[0.5,5.5]$, by which an integer in $\{1,2,3,4,5\}$ can be assigned as the dilation factor for each point. To summarize, the dilation factors learning in our implementation follows:
$$
    \mathcal{D}_{{N}} = \left\lceil 5\cdot{\frac{1}{1+e^{\left({Conv_{[1,1]}^{\big\{({k\cdot d_{max}}/2),1\big\}}}({E}_{{N}\times{(k\cdot d_{max})}})\right)}}} + 0.5 \right\rfloor
$$

\begin{table}
\begin{center}
\resizebox{0.95\columnwidth}{!}{
\begin{tabular}{l|ccc}
\Xhline{3\arrayrulewidth}
method    & model size (MB) & time (ms) & overall acc. (\%) \\\hline
PointNet \cite{qi2017pointnet}        & 40 & 16.6 & 89.2         \\
PointNet++ \cite{qi2017pointnet++}        & 12 &163.2 & 90.7      \\
PCNN \cite{atzmon2018point}        & 94  &117.0     & 92.3      \\
DGCNN \cite{wang2019dynamic}        & 21 &27.2 & 92.9  \\
\textbf{Ours}       & 70 &$19.2^\star$ & \textbf{93.1}  \\\Xhline{3\arrayrulewidth}
\end{tabular}
}
\end{center}
\caption{Complexity of classification network on \emph{ModelNet40}. ($^*$running on GeForce GTX 2080Ti)}
\label{tab:complexity}
\end{table}
\subsection{Loss Function}
As discussed in the Section 3.3, the total loss for training is the sum of cross-entropy loss $\mathcal{L}_{ce}$ and weighted error-minimizing module losses: $\sum{w_i\cdot \mathcal{L}_{{er}_{i}}}$. In practice, we apply 4 error-minimizing modules in the Full-resolution (FR) branch of our network, adopting the similar layers and feature dimensions as in~\cite{wang2019dynamic}. In terms of our experiments on the ShapeNet Part, ModelNet40 and ScanObjectNN datasets, we empirically set a larger weight for the first error-minimizing module ($w_1 = 0.1$) since its output affects the both branches and constrains the network learning at the beginning. In contrast, the weights for other modules' losses can be smaller ($w_2 = w_3 = w_4 = 0.01$). Although the additional losses are incorporated, the cross-entropy loss still contributes the most to the training. The overall loss $\mathcal{L}$ in our practice is formulated as:
$$
    \mathcal{L} = \mathcal{L}_{ce} + {0.1\cdot \mathcal{L}_{{er}_{1}}} + {0.01\cdot \mathcal{L}_{{er}_{2}}} + {0.01\cdot \mathcal{L}_{{er}_{3}}} + {0.01\cdot \mathcal{L}_{{er}_{4}}}.
$$

\section{Multi-resolution Branch}
\begin{figure*}
\begin{center}
\includegraphics[width=0.8\textwidth]{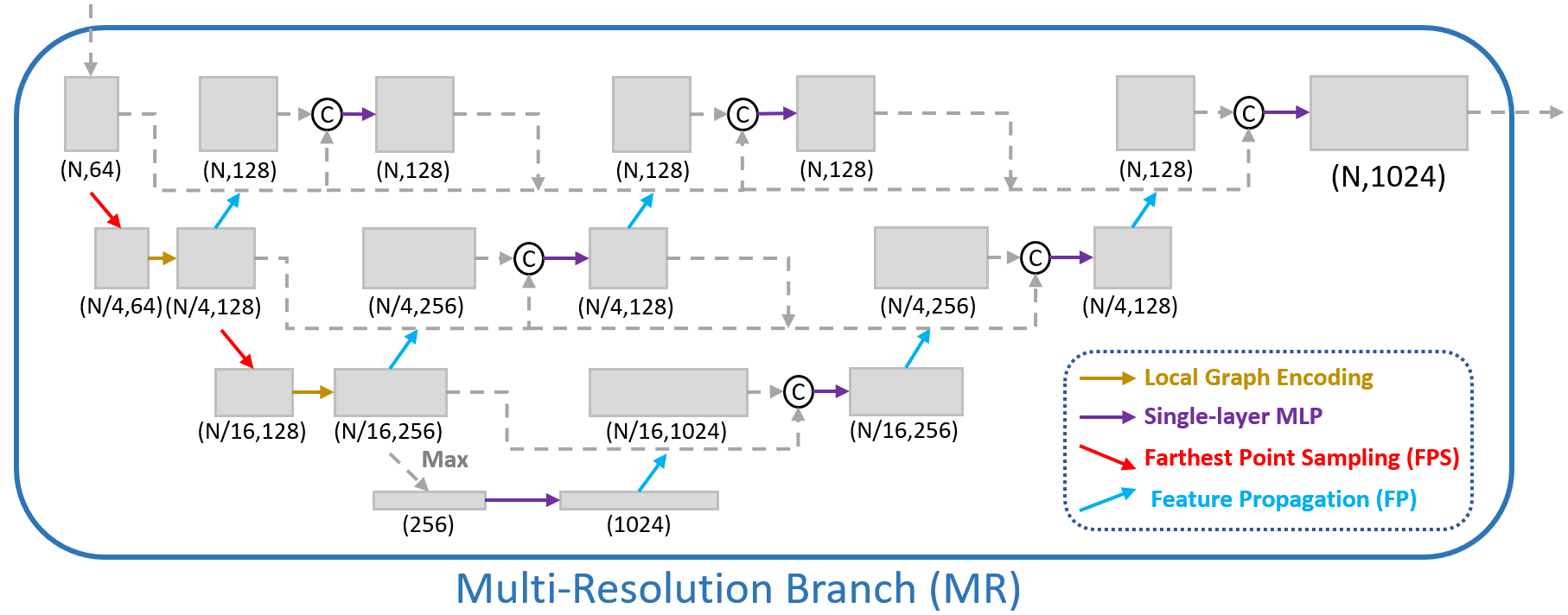}
\end{center}
\caption{The input of the MR branch is the output of the first error-minimizing module in the FR branch, while the output of the MR branch merges with the output of the FR branch following the behavior as Equation 4 in the main paper.}
\label{fig:mr_details}
\end{figure*}

As shown in Figure~\ref{fig:mr_details}, the MR branch is implemented with light-weight operations such as single-layer MLPs, and only investigates 2 more resolutions of the point cloud using basic Local Graph Encoding as Equation 2 in the main paper. For upsampling and downsampling operations, they are implemented based on CUDA without learnable weights. Besides, we use the dense connections and concatenations to enhance the relations between the feature maps of different resolutions.

\section{Model Complexity}
In addition, we adopt the network complexity data provided in~\cite{wang2019dynamic} for a fair comparison. As Table~\ref{tab:complexity} shows, our model size is relatively large due to the parameters and operations needed. However, the inference time of our method running on a single GeForce GTX 2080Ti GPU is only \emph{19.2 ms}, which indicates the ability of our model in forward procedure thanks to the algorithm optimization and relevant CUDA implementation.

\end{document}